\documentclass{article}
\PassOptionsToPackage{numbers, compress}{natbib}
\usepackage[preprint]{neurips_2025}
\usepackage{amssymb, pifont, textcomp}
\usepackage{pifont}
\usepackage{xcolor}
\usepackage{tabularray}
\usepackage{booktabs}
\usepackage{pifont}
\usepackage{array}
\usepackage{multirow}
\usepackage{graphicx}
\usepackage{caption}
\usepackage{subcaption}
\usepackage{tabularx}
\usepackage{wrapfig} % 必需宏包
\usepackage{booktabs}
\usepackage{makecell}

\usepackage[utf8]{inputenc} % allow utf-8 input
\usepackage[T1]{fontenc}    % use 8-bit T1 fonts
\usepackage{hyperref}       % hyperlinks
\usepackage{url}            % simple URL typesetting
\usepackage{booktabs}       % professional-quality tables
\usepackage{amsfonts}       % blackboard math symbols
\usepackage{nicefrac}       % compact symbols for 1/2, etc.
\usepackage{microtype}      % microtypography
\usepackage{xcolor}         % colors
\usepackage{hyperref} % 引入 hyperref 宏包

\usepackage{graphicx}
\usepackage{amsmath}
\usepackage{booktabs}
\usepackage{longtable}
\usepackage{hyperref}
\usepackage{geometry}
\usepackage{booktabs}
\title{RealTalk-CN: A Realistic Chinese Speech-Text Dialogue Benchmark With Cross-Modal Interaction Analysis}

\author{
    Enzhi Wang$^{1}$, Qicheng Li$^{1}$, Shiwan Zhao$^{1}$, Aobo Kong$^{1}$, Jiaming Zhou$^{1}$,
    \\ Xi Yang$^{2}$, Yequan Wang$^{2}$, Yonghua Lin$^{2}$, Yong Qin$^{1*}$ \\
    $^1$TMCC, College of Computer Science, Nankai University, \\
    $^2$Beijing Academy of Artificial Intelligence (BAAI), Beijing, China \\
    Correspondence: \texttt{liqicheng@mail.nankai.edu.cn}, \texttt{qinyong@nankai.edu.cn}
}

\begin{document}

\maketitle

\begin{abstract}
In recent years, large language models (LLMs) have achieved remarkable advancements in multimodal processing, including end-to-end speech-based language models that enable natural interactions and perform specific tasks in task-oriented dialogue (TOD) systems. However, existing TOD datasets are predominantly text-based, lacking real speech signals that are essential for evaluating the robustness of speech-based LLMs. Moreover, existing speech TOD datasets are primarily English and lack critical aspects such as speech disfluencies and speaker variations. To address these gaps, we introduce \textbf{RealTalk-CN}, the first Chinese multi-turn, multi-domain speech-text dual-modal TOD dataset, comprising 5.4k dialogues (60K utterances, 150 hours) with paired speech-text annotations. RealTalk-CN captures diverse dialogue scenarios with annotated spontaneous speech disfluencies, ensuring comprehensive coverage of real-world complexities in speech dialogue. In addition, we propose a novel cross-modal chat task that authentically simulates real-world user interactions, allowing dynamic switching between speech and text modalities. Our evaluation covers robustness to speech disfluencies, sensitivity to speaker characteristics, and cross-domain performance. Extensive experiments validate RealTalk-CN’s effectiveness, establishing a strong foundation for Chinese speech-based LLMs research. Data and code will be available.
\end{abstract}

\section{Introduction}
In recent years, large language models (LLMs) have achieved significant breakthroughs in multimodal processing. For speech input, this has led to the emergence of end-to-end speech-based language models (e.g., GPT-4o \cite{achiam2023gpt4}, Qwen2-Audio~\cite{chu2024qwen2audio} and Baichuan-Omni~\cite{li2024baichuanomni}). These models can directly conduct end-to-end speech conversations with humans and perform specific tasks, improving the efficiency, naturalness, and authenticity of the conversations. Therefore, traditional text-based task-oriented dialogue (TOD) systems~\cite{cai-etal-2024-unipcm} that help users achieve specific goals have also been transformed into speech interaction forms.

However, existing task-oriented dialogue datasets are predominantly text-based, such as the large-scale English dataset MultiWOZ~\cite{zang2020multiwoz} or Chinese datasets like CrossWOZ~\cite{zhu2020crosswoz} and RiSAWOZ~\cite{quan2020risawoz}. These lack real speech signals, making it difficult to evaluate models’ robustness against spontaneous speech disfluencies phenomena (e.g., fillers, hesitations, repetitions and self-corrections)~\cite{shriberg1994disfluency} or speaker variations (gender/age/regional accent)~\cite{krause2004clear}. This severely limits their applicability for evaluating speech-based LLMs.

For speech task-oriented dialogue data, resources remain scarce. Early English datasets such as ATIS~\cite{hemphill1990atis} and DSTC2/10~\cite{henderson2014dstc2, kim2021robustness} are single-turn speech understanding datasets with limited evaluation scenarios. The recent SpokenWOZ~\cite{si2023spokenwoz} advances the field as the first large-scale multi-turn, multi-domain English speech TOD dataset, providing a new benchmark. However, it does not annotate speech signals with disfluent speech, which is crucial for evaluating the robustness of speech-based LLMs in TOD scenarios. In addition, there is no similar Speech TOD dataset for Chinese scenarios, while Chinese spoken dialogue exhibits unique linguistic phenomena and sociocultural characteristics~\cite{huang2023ceval}, and China has seen rapid development of powerful speech-based and even omnimodal LLMs (e.g., Baichuan-Audio~\cite{li2025baichuanaudio}, GLM-4-Voice~\cite{zeng2024glm4voice}, Qwen-2.5-Omni~\cite{xu2025qwen25}). This creates a growing gap between model development and evaluation in Chinese speech-based LLMs.

In addition, while previous discussions focused on speech-only interactions, real-world dialogues often involve modality switching between speech and text. Current studies on speech-text multimodal TOD systems~\cite{si2023spokenwoz, li2024multimodal} mainly assume that users provide speech and text inputs at the same time, combining both modalities to improve responses. However, this approach does not match how people actually interact with voice assistants, where they naturally alternate between speaking and typing across different turns—for instance, asking a voice question about product details and then typing a follow-up text request for discounts in the applications of speech LLMs like e-commerce AI assistants. This common usage pattern reveals a key limitation in existing evaluations: current benchmarks cannot test whether models properly handle such dynamic modality switching. To better assess the real-world performance of speech-based LLMs in TOD system applications, we need task designs that reflect natural modality changes during conversations.
To address these problems, we propose:
\begin{itemize}
    \item \textbf{A novel dataset}: The first Chinese multi-turn, multi-domain speech-text dual-modal TOD dataset, RealTalk-CN, comprising 5.4k dialogues (60K utterances, 150 hours) with paired speech-text annotations. It labels spontaneous speech disfluencies phenomena, covering a wide range of domains and diverse speakers to reflect real-world dialogue complexities. All data are real human-to-human speech conversations.
    
    \item \textbf{Innovative Task}: A novel cross-modal chat task, where users and assistants can dynamically alternate between speech and text modalities during conversations (e.g., voice queries followed by text-based feedback via SMS or apps). This design authentically simulates real-world conversations between humans and assistants, enabling comprehensive evaluation of models' cross-modal information integration and context management capabilities.
    
    \item \textbf{Comprehensive Evaluation}: A systematic experimental protocol across four dataset subsets, incorporating diverse open-source and proprietary baseline models. The evaluation spans three critical dimensions: (i) robustness to speech disfluencies (e.g., Modal particle drag, hesitations, repetitions, self-corrections) in TOD scenarios, (ii) sensitivity analysis to speaker characteristics (gender/age/region), and (iii) cross-domain performance comparison. We employ GPT-4-based automated evaluation to ensure consistent and scalable metrics.
    
    \item \textbf{Rigorous Data Quality Control}: The data collection and annotation process incorporates multiple quality assurance measures, including standardized script design, controlled recording environments, balanced speaker diversity representation, detailed annotation guidelines, and stringent management protocols. These procedures collectively ensure the ecological validity and reliability of the dataset for research purposes.
\end{itemize}

\section{Related Work}
Table~\ref{related_work} summarizes the various aspects of our dataset compared with other related datasets. Related work can be roughly divided into three categories:

\textbf{Text-based TOD datasets}: English resources in this domain include MultiWOZ~\cite{zang2020multiwoz}, a widely used dataset spanning eight domains with over ten thousand dialogues. For Chinese, notable datasets are CrossWOZ~\cite{zhu2020crosswoz}, which contains six thousand dialogues and 102 thousand utterances, and RiSAWOZ~\cite{quan2020risawoz}, a more extensive collection featuring 11.2 thousand dialogues, 150 thousand utterances, and coverage across twelve domains. These provide rich annotations for dialogue state tracking but lack speech signals.

\textbf{Spoken language understanding (SLU) datasets}: Most English SLU datasets such as SNIPS~\cite{kawar2021snips} rely on transcribed text without accounting for speech recognition errors. The largest existing English SLU resource is SLURP~\cite{bastianelli2020slurp}, which covers eighteen domains. In contrast, Chinese research has seen initial progress with CATSLU~\cite{zhu2019catslu}, a multi-domain audio-text dataset introduced during the ICMI 2019 challenge. However, these datasets are only single-turn content understanding tasks.

\textbf{Speech-based TOD datasets}: Existing speech-based task-oriented datasets remain scarce. Early efforts such as DSTC2~\cite{henderson2014dstc2} and DSTC10~\cite{kim2021robustness} provide only small-scale automatic speech recognition outputs. SpokenWOZ~\cite{si2023spokenwoz} represents the first large-scale English speech-text benchmark but lacks speech disfluency annotation and speaker feature annotation. Moreover, no similarly comprehensive Chinese dataset currently exists, creating a significant gap that impedes research progress in this area.
% SpokenWOZ~\cite{si2023spokenwoz} represents the first large-scale English speech-text benchmark. However, no similarly comprehensive Chinese dataset currently exists, creating a significant gap that impedes research progress in this area.
\begin{table}
\centering
\tiny
\caption{Comparison of our dataset with other related datasets. TOD stands for Task-Oriented Dialogue Dataset, SLU is a single-round Spoken Language Understanding dataset. H2H, H2M, M2M stand for human-to-human, human-to-machine, machine-to-machine.}
\begin{tblr}{
  width = \linewidth,
  colspec = {Q[129]Q[133]Q[77]Q[75]Q[79]Q[81]Q[75]Q[48]Q[58]Q[87]Q[94]},
  column{3} = {c},
  column{4} = {c},
  column{5} = {c},
  column{6} = {c},
  column{7} = {c},
  column{8} = {c},
  column{9} = {c},
  column{10} = {c},
  column{11} = {c},
  cell{2}{1} = {r=6}{},
  cell{8}{1} = {r=4}{},
  cell{12}{1} = {r=3}{},
  % vline{1} = {1-15}{0.05em}, % 添加最左侧竖线
  vline{2} = {1-15}{0.05em}, % 强化第一列后的竖线
  vline{3} = {1-15}{0.05em}, % 确保所有竖线完整
  hline{1,15} = {-}{0.08em},
  hline{2,8,12} = {-}{0.05em},
  rowsep = 1pt, % 减小行间距
  colsep = 1pt, % 减小列间距
}
Type             & Dataset                  & Language & Speakers & Dialogues & Avg. turns & Domains & Slots & Audio & {Disfluency\\Annotation} & {Cross modal \\task} \\
\hline
Text-based TOD   & \makecell[l]{\textbf{M2M}~\cite{shah2018selfplay}}      & EN       & M2M      & 1,500     & 9.9        & 2       & 14    & \textbf{\ding{56}}    & \textbf{\ding{56}}                       & \textbf{\ding{56}}                   \\
                 & \makecell[l]{\textbf{KVRET}~\cite{eric2017kvret}}     & EN       & H2H      & 2,425     & 5.3        & 3       & 13    & \textbf{\ding{56}}    & \textbf{\ding{56}}                       & \textbf{\ding{56}}                   \\
                 & \makecell[l]{\textbf{MultiWOZ}~\cite{budzianowski2018multiwoz}}  & EN       & H2H      & 8,438     & 13.7       & 7       & 25    & \textbf{\ding{56}}    & \textbf{\ding{56}}                       & \textbf{\ding{56}}                   \\
                 & \makecell[l]{\textbf{DSTC10}~\cite{kim2021robustness}}    & EN       & H2H      & 107       & 21.4       & 3       & -     & \textbf{\ding{56}}    & \textbf{\ding{56}}                       & \textbf{\ding{56}}                   \\
                 & \makecell[l]{\textbf{CrossWOZ}~\cite{zhu2020crosswoz}}   & ZH       & H2H      & 5,012     & 16.9       & 5       & 72    & \textbf{\ding{56}}    & \textbf{\ding{56}}                       & \textbf{\ding{56}}                   \\
                 & \makecell[l]{\textbf{RISAWOZ}~\cite{quan2020risawoz}}    & ZH       & H2H      & 10,000    & 13.5       & 12      & 159   & \textbf{\ding{56}}    & \textbf{\ding{56}}                       & \textbf{\ding{56}}                   \\
\hline
Speech-based SLU & \makecell[l]{\textbf{FSC}~\cite{lugosch2019pretraining}}       & EN       & H        & 30,043    & 1          & 1       & -     & \textbf{\ding{51}}   & \textbf{\ding{56}}                       & \textbf{\ding{56}}                   \\
                 & \makecell[l]{\textbf{SNIPS}~\cite{kawar2021snips}}     & EN       & H        & 13,084    & 1          & 7       & 72    & \textbf{\ding{51}}   & \textbf{\ding{56}}                       & \textbf{\ding{56}}                   \\
                 & \makecell[l]{\textbf{SLURP}~\cite{bastianelli2020slurp}}     & EN       & H        & 72,277    & 1          & 18      & 55    & \textbf{\ding{51}}   & \textbf{\ding{56}}                       & \textbf{\ding{56}}                   \\
                 & \makecell[l]{\textbf{CATSLU}~\cite{zhu2019catslu}}    & ZH       & H        & 16,258    & 1          & 4       & 94    & \textbf{\ding{51}}   & \textbf{\ding{56}}                       & \textbf{\ding{56}}                   \\
\hline
Speech-based TOD & \makecell[l]{\textbf{DSTC2}~\cite{henderson2014dstc2}}     & EN       & H2M      & 1,612     & 14.5       & 1       & 8     & \textbf{\ding{51}}   & \textbf{\ding{56}}                       & \textbf{\ding{56}}                   \\
                 & \makecell[l]{\textbf{SpokenWOZ}~\cite{si2023spokenwoz}} & EN       & H2H      & 5,700     & 35.5       & 26      & 36    & \textbf{\ding{51}}   & \textbf{\ding{56}}                       & \textbf{\ding{56}}                   \\
                 & \textbf{RealTalk-CN}(ours) & \textbf{ZH }      & H2H      & 5,400     & 12.1       & 58      & 115   & \textbf{\ding{51}}   & \textbf{\ding{51}}                      & \textbf{\ding{51}}                  
\end{tblr}
\label{related_work}
\end{table}
\section{RealTalk-CN Data Collection and Quality Control}
During the data collection phase, we prioritized speech quality and annotation consistency. The dataset was constructed using pre-written scripts designed to reflect natural spoken language characteristics, including casual grammar, colloquial vocabulary, short sentence structures, and loose syntactic organization~\cite{mccarthy1995spoken}. The dialogues covered multiple domains while allowing participants to improvise on the recording to maintain conversational authenticity. Crucially, 10\% of the collected data intentionally preserved spontaneous speech disfluencies such as repetitions, hesitations, self-corrections, and modal particle drag to simulate real-world conditions.

For speech-text alignment, we implemented a rigorous timestamping mechanism to mark utterance boundaries and dialogue turns. Recording sessions were conducted in quiet indoor environments using both professional microphones and consumer-grade smartphone microphones to ensure device diversity representative of real usage scenarios. Dual recording methods (dedicated recorders and smartphones) were employed to capture authentic acoustic conditions.

\begin{figure}[htbp]
\centering
\includegraphics[width=0.95\textwidth]{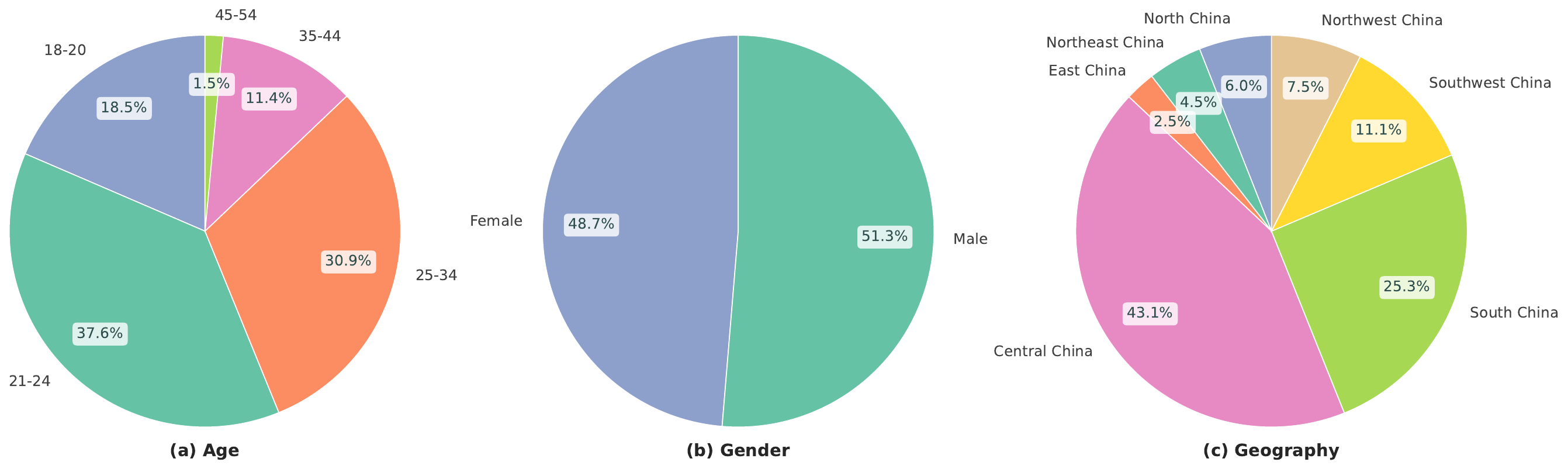}
\caption{The distribution of Speakers. The dataset covers most age groups. It also has a near-equal gender split. It covers all major regions in China to explore the impact of different regional accents on the speech model. The specific provinces included in each region are in the appendix A.1.}
\label{fig:full_dist}
\end{figure}

Speaker diversity was ensured through 300 volunteers (gender ratio 1:1±10\%, ages 18-50 following normal distribution covering young and middle-aged demographics) with predominantly Mandarin proficiency while permitting mild regional accents. The gender, age, and regional distribution are shown in Figure~\ref{fig:full_dist}. Each participant contributed to 50 dialogue sessions.% yielding 150 hours of validated speech data.

Annotation consistency was maintained through multi-round verification with detailed guidelines addressing various Chinese speech phenomena. Transcripts were required to strictly match actual pronunciations while accommodating dialectal variations, such as converting "Liu nai" to standard "Niu nai". Mandarin phonological features including erhua were preserved in transcriptions. %English terms followed case-sensitive conventions, distinguishing between proper nouns ("iPhone") and spelled-out letters ("Q Q"). 
Standard references were used to verify proper nouns, while numerical expressions were consistently rendered in Chinese characters. Filler sounds and discourse markers were retained to maintain prosodic authenticity, with special notation applied to the intentionally preserved 10\% of data containing disfluencies.
Comprehensive quality control measures were implemented throughout the process. Audio clips maintained 0.2-0.3 seconds of silence padding with duration optimized at 5-6 seconds (maximum 12 seconds). A 5\% random sampling protocol ensured slot-value annotation accuracy exceeded 95\%. The annotation Pipeline incorporated iterative optimization, beginning with pilot annotation of three sample batches to refine guidelines before full-scale implementation. %Data was delivered in progressive 10\% increments with rigorous quality reviews at each stage.
The Ethics Statement of the dataset is described in Appendix D.1.

\section{RealTalk-CN Dataset Overview}
RealTalk-CN represents the first Chinese multi-turn, multi-domain speech-text dual-modal TOD dataset, which comprises 5.4k dialogue sessions, including 1.2k single-domain and 4.2k cross-domain conversations, totaling over 60k utterances contributed by 113 speakers. With an average of 12.1 turns per dialogue and 150 hours of validated audio, the dataset covers dozens of task-oriented domains (e.g., dining, transportation, shopping) through authentic human-to-human interactions. Each dialogue is accompanied by comprehensive annotations including dialogue states (slots), intents, transcriptions, utterance-level timestamps, speaker metadata, and labels for spontaneous speech disfluencies phenomena (e.g., filled pauses, repetitions, self-corrections). %These annotations support multiple research tasks such as intent classification, dialogue state tracking, and response generation.

\subsection{Spontaneous Speech Phenomena}
\begin{wrapfigure}{r}{0.5\textwidth}
    \centering
    \includegraphics[width=0.9\linewidth]{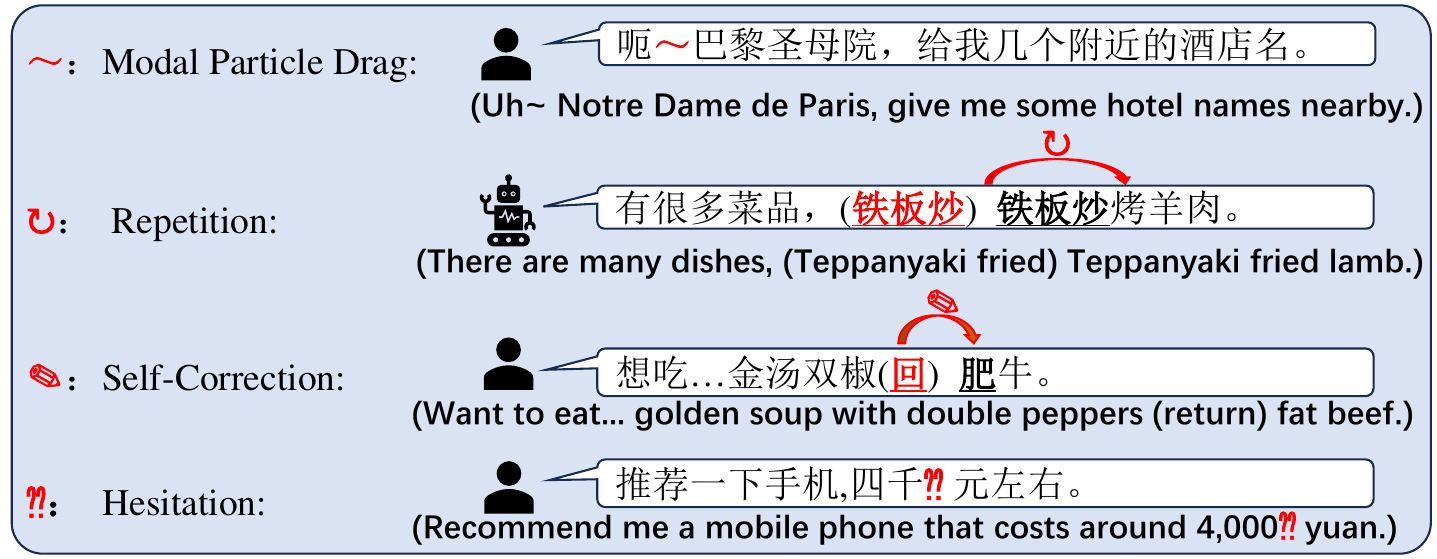}
    \caption{Four speech disfluency types in Chinese dialogues: Modal particle drag, repetition, self-correction, and hesitation. The third example is that in Chinese, "return" and "fat" sound similar.}
    \label{fig:disfluency_example}
\end{wrapfigure}
As a spoken language-oriented resource, RealTalk-CN captures fundamental distinctions between oral and written communication styles - even within identical semantic contexts, spoken dialogues exhibit casual grammar, colloquial vocabulary, fragmented structures, and loose syntactic organization~\cite{mccarthy1995spoken}. Our scripting process explicitly mandated conversational language patterns. Crucially, spoken disfluencies~\cite{shriberg1994disfluency} present additional challenges for language understanding systems. While the English SpokenWOZ dataset~\cite{si2023spokenwoz} addressed this partially through ASR-derived noise, it lacked explicit annotation of disfluency types. RealTalk-CN advances this through systematic labeling of spontaneous speech disfluencies, enabling the creation of phenomenon-specific subsets for robustness evaluations of speech-based LLMs. This design also facilitates secondary applications like speech disfluency correction. As illustrated in Figure \ref{fig:disfluency_example}, we defined common disfluency categories, instructed speakers to maintain natural conversational flow (including organic production of disfluencies), and implemented rigorous post-hoc annotation protocols.

\subsection{Broad Domain Coverage}
\begin{table}
\centering
\tiny
\caption{Data statistics of the four subsets. Colloquial means that the text content contains the above-mentioned unfluent spoken language markers, while System means the opposite. and Avg Intent Choices means the average number of candidate intents as answers for each utterance. MD means Multi-Domain, and SD means Single-Domain. M, R, S, and H respectively represent Modal Particle Drag, Repetition, Self-Correction, and Hesitation}
\label{subset_tables}
\small
\begin{tblr}{
  width = \linewidth,
  colspec = {Q[120]Q[75]Q[170]Q[150]Q[145]Q[180]}, % 各列宽度略微减小
  cells = {c},
  hline{1,6} = {-}{0.08em},
  hline{2} = {-}{0.05em},
  hline{4} = {-}{dashed},
  colsep = 1.35pt, % 列间距减小
  stretch = -1, % 略微压缩单元格内边距
}
Subsets        & Samples & Avg Utterance Length & Avg Dialog Rounds & Avg Intent Choices & Avg Disfluency Markers                   \\
MD-Col & 3,837   & 27.42                & 8.54              & 34.51              & {M: 0.12\quad R: 0.04\\ S: 0.11\quad H: 1.14} \\
MD-Sys     & 3,837   & 19.27                & 7.73              & 34.77              & {-} \\
SD-Col & 892     & 25.61                & 8.14              & 25.90               & {M: 0.63\quad R: 0.07\\S: 0.18\quad H: 0.52} \\
SD-Sys     & 892     & 20.76                & 7.58              & 27.03              & {-} 
\end{tblr}
\end{table}

\begin{figure}[htbp]
    \centering
    \includegraphics[width=0.63\textwidth]{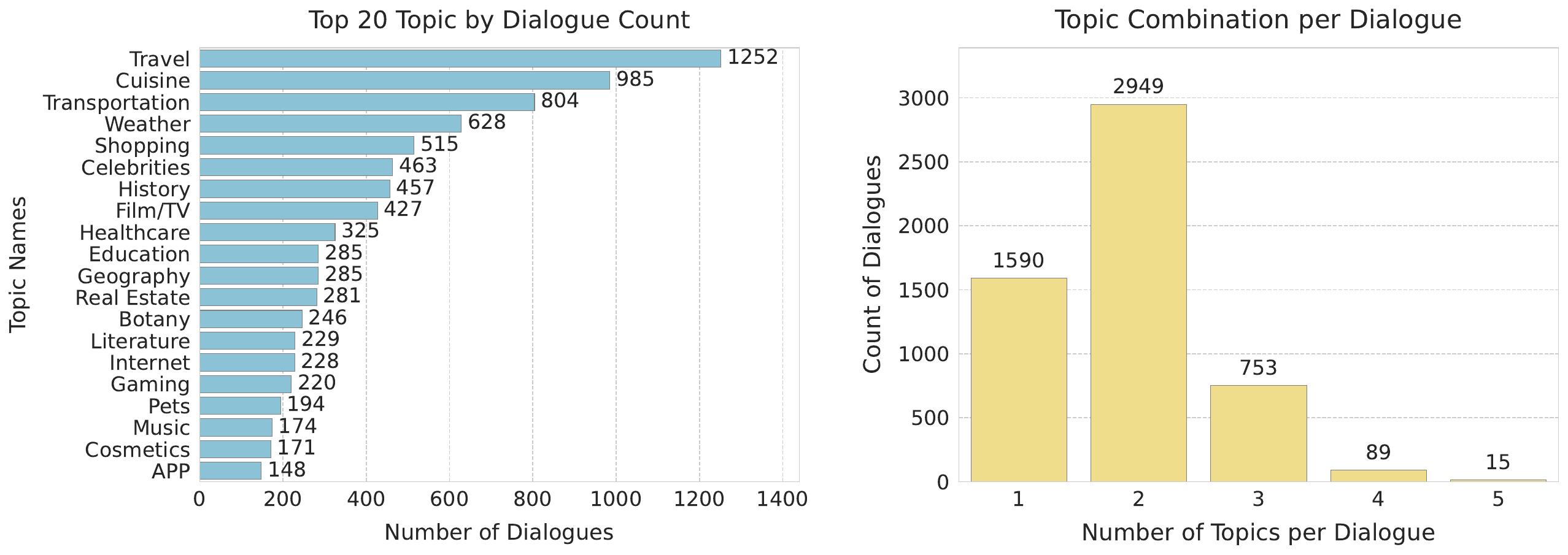}
    \caption{Domain distribution and multi-domain combination distribution of RealTalk-CN.}
    \label{fig:topic_distribution_dialogues}
\end{figure}

RealTalk-CN comprehensively encompasses 58 TOD domains, including weather, dining, travel, news, shopping, finance, and healthcare. It also has 55 intents and 115 slot types, which are not available in previous datasets. Detailed intent and slot information can be found in Appendix A.2. The dataset is systematically organized into single-domain and multi-domain dialogues, with the latter involving 2-5 interleaved domains to better simulate real-world scenarios. For example, a travel-related conversation can naturally incorporate weather inquiries and restaurant recommendations. As illustrated in Figure \ref{fig:topic_distribution_dialogues}, the domain distribution follows a long-tail pattern: high-frequency domains (e.g., travel, weather) cover common daily topics, while mid-to-low frequency domains ensure comprehensive topical diversity. Among multi-domain dialogues, 2,949 sessions involve two domains (representing the majority), followed by 753 sessions with three domains. Additionally, the data set includes complex dialogues that span 4-5 domains.

% 基本图片插入

To enable granular analysis, we partition the dataset into four subsets based on two criteria: (1) single-domain vs. multi-domain composition, and (2) presence of annotated speech disfluencies. Detailed statistics of the test set are provided in Table \ref{subset_tables}. The detailed data of the training, validation, and test set divisions can be found in Appendix A.3. Multi-domain dialogues demonstrate significantly greater complexity than single-domain counterparts, evidenced by higher average intent counts, reflecting more diverse user needs in cross-domain interactions. Linguistically, disfluency-annotated subsets exhibit longer utterance lengths and more dialogue turns due to phenomena such as self-corrections, repetitions, and Modal particle drags. These characteristics mirror authentic speech patterns and present increased robustness challenges for speech-based LLMs in TOD scenarios.

\subsection{Innovative Cross-Modal Chat Task}
\begin{figure}[htbp]
    \centering
    \includegraphics[width=0.50\textwidth]{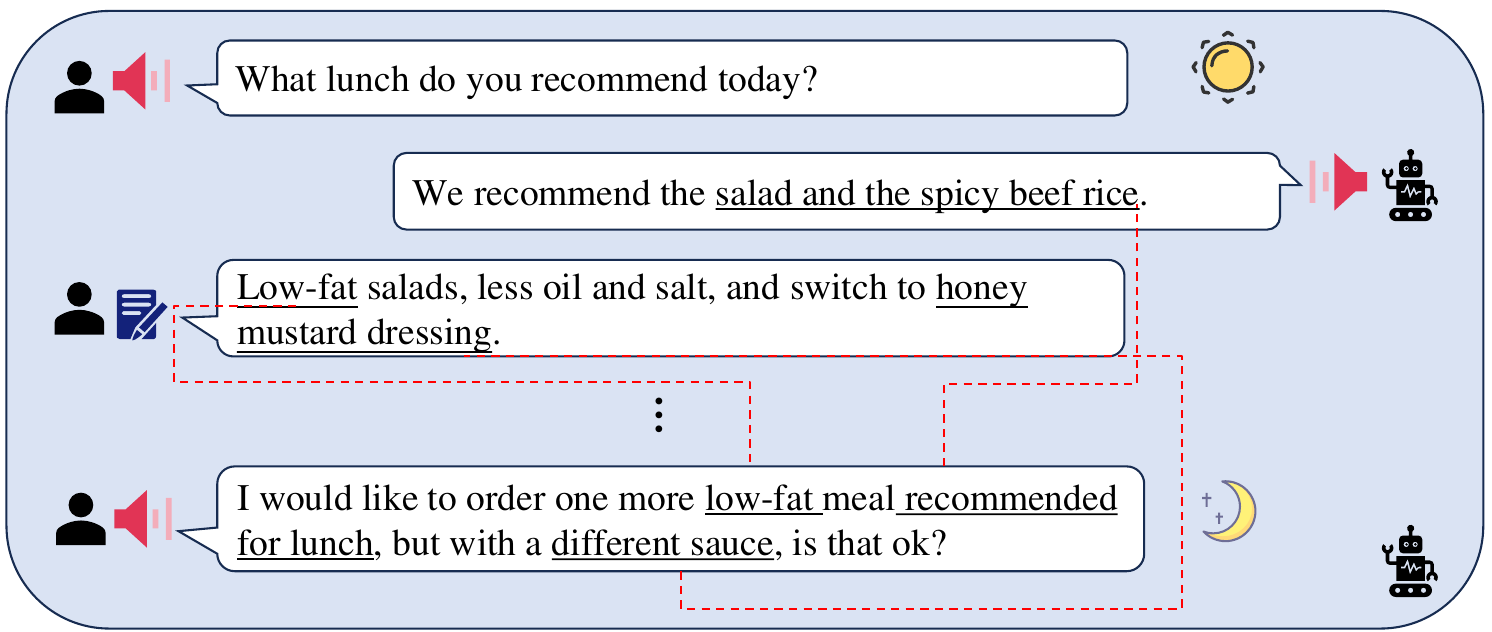}
    \caption{An example of dynamically switching speech-text modality dialogue. In this scenario, the system must integrate the dish recommendations provided via noon voice messages with the user's customized preferences expressed in text, accurately comprehend that "low-fat meal" refers to salad, and correctly identify the user's request to change the dressing.} %, thereby providing dish options that align with the user's health preferences.}
    \label{fig:mix_modality_example}
\end{figure}

Current research on multimodal dialogue systems focuses primarily on scenarios in which users and systems simultaneously receive and process multiple modalities, such as speech and text. For example, previous work~\cite{si2023spokenwoz}~\cite{li2024multimodal} proposed multimodal speech-text dialogue datasets where the evaluation task involves responding to contexts that contain speech and text modalities, aiming to enhance textual representations through aggregated speech embeddings for improved responses. However, this simultaneous multimodal input paradigm rarely occurs in real-world applications. In practical intelligent voice assistant usage, user-system interactions typically span multiple turns with dynamic modality switching. For example, in a restaurant reservation scenario, users might initially inquire via voice and subsequently continue the conversation through text messages or mobile apps, rather than providing identical content through both speech and text simultaneously. A concrete illustration of this pattern is shown in Figure \ref{fig:mix_modality_example}.

To address this gap, we propose a novel cross-modal chat task where the conversational context contains mixed speech or text utterances without simultaneous modality presentation. The key challenge lies in the model's ability to accurately comprehend and track information distributed across different modalities while effectively integrating these heterogeneous inputs to generate consistent and coherent responses. To isolate the impact of modality switching from speech disfluency effects, we specifically employ speech modality for turns containing any of the four disfluency markers, while using text modality otherwise, thereby creating a dynamically switching context.
\section{Experimental Design \& Evaluation}
\subsection{Task Design}
We designed multiple tasks on the SpokenMMC dataset to fully exploit its potential. We use the same zero-shot evaluation protocol for speech-based LLMs following~\cite{yang2024airbench, chen2024voicebench}, and we also provide the training set for researchers to use.

\textbf{Standard task-oriented dialogue tasks}, including dialogue intent classification, slot filling, and end-to-end chat. Following~\cite{chen2024voicebench}, for the intent classification task, we compute Accuracy and the PANDA discriminant~\cite{li2024panda} estimation method which has a strong correlation with human evaluation. For the slot filling task, we use the classic evaluation metrics F1 and joint goal accuracy (JGA)~\cite{budzianowski2018multiwoz}. Specific examples of the tasks and our evaluation process can be found in Appendix B.1. For the chat task, since traditional metrics have demonstrated a weak correlation with human judgment~\cite{liu2023gpteval}, we implemented GPT-4-based automatic evaluation following~\cite{chen2024voicebench}~\cite{liu2023gpteval}~\cite{yang2024airbench}. All evaluations are conducted using GPT-4o-mini\footnote{GPT-4o-mini-2024-07-18}, including the sum of the scores of the evaluation without reference and the evaluation with reference. The evaluation prompts can be found in Appendix C.1

\textbf{Cross-modal chat task}, as described earlier, users and assistants dynamically switch between speech and text modalities during conversations.

\textbf{Robustness evaluation task} leverages the annotated speech disfluencies to examine models' tolerance to conversational incoherence, using performance differences on the Colloquial subset as the evaluation metric.

\subsection{Baselines}
We evaluated several end-to-end speech-based LLMs, including Qwen2-Audio-7B-Instruct~\cite{chu2024qwen2audio}, Baichuan-Audio-Instruct~\cite{li2025baichuanaudio}, GLM-4-Voice-9B~\cite{zeng2024glm4voice}, along with recent Omni-modal foundation models (MiniCPM-o~\cite{yao2024minicpmv}, Baichuan-Omni-1d5~\cite{li2024baichuanomni}, Qwen2.5-Omni-7B~\cite{xu2025qwen25}). For comparison, we also included \textbf{Pipeline} approaches combining Whisper-Large-V3~\cite{radford2023whisper} with text-only LLMs (GPT-4o) and GPT-4o-Audio-mini\footnote{GPT-4o-mini-audio-preview}, aiming to measure performance gaps between current open/closed-source voice LLMs and traditional Pipeline methods. We evaluated the models based on the code in~\cite{chen2024voicebench}.

\begin{table}
\centering
\scriptsize
\caption{Performance comparison of the models on the intent classification (IC) and slot filling (SF) tasks of the RealTalk-CN dataset. Acc is the accuracy of intent classification, and Pipeline represents Whisper-large-v3 + GPT-4o. GPT-4o-Audio uses the mini version. PAN. represents PANDA score.}
\label{table1}
\begin{tblr}{
  width = \linewidth,
  colspec = {Q[100]Q[38]Q[42]Q[38]Q[38]Q[38]Q[42]Q[38]Q[38]Q[38]Q[42]Q[38]Q[38]Q[38]Q[42]Q[38]Q[38]Q[44]}, % 第一列从78增加到100，其他列相应减小
  column{even} = {c},
  column{3} = {c},
  column{5} = {c},
  column{7} = {c},
  column{9} = {c},
  column{11} = {c},
  column{13} = {c},
  column{15} = {c},
  column{17} = {c},
  cell{1}{2} = {c=4}{0.16\linewidth},
  cell{1}{6} = {c=4}{0.16\linewidth},
  cell{1}{10} = {c=4}{0.16\linewidth},
  cell{1}{14} = {c=4}{0.16\linewidth},
  cell{2}{2} = {c=2}{0.08\linewidth},
  cell{2}{4} = {c=2}{0.08\linewidth},
  cell{2}{6} = {c=2}{0.08\linewidth},
  cell{2}{8} = {c=2}{0.08\linewidth},
  cell{2}{10} = {c=2}{0.08\linewidth},
  cell{2}{12} = {c=2}{0.08\linewidth},
  cell{2}{14} = {c=2}{0.08\linewidth},
  cell{2}{16} = {c=2}{0.08\linewidth},
  vline{2} = {1-2}{},
  vline{4,8,12,16} = {2-11}{dashed},
  vline{2,6,10,14} = {3-11}{},
  hline{1} = {-}{0.08em},
  hline{2-5,11-12} = {1-17}{},
  hline{2-5,11-12} = {18}{0.03em},
  rowsep = 0.5pt,
  colsep = 1.5pt}
Subsets           & MD-Col         &                &                &                & MD-Sys         &                &                &                & SD-Col         &                &                &                & SD-Sys         &                &                &                &         \\
Tasks             & IC             &                & SF             &                & IC             &                & SF             &                & IC             &                & SF             &                & IC             &                & SF             &                &         \\
Metrics           & Acc            & PAN.          & F1             & JGA            & Acc            & PAN.          & F1             & JGA            & Acc            & PAN.          & F1             & JGA            & Acc            & PAN.          & F1             & JGA            & Average \\
Pipeline           & \textbf{53.56} & \textbf{53.56} & 45.90          & 26.09          & \textbf{54.83} & \textbf{54.83} & 48.81          & 31.99          & \textbf{59.75} & \textbf{59.75} & 38.55          & 20.68          & \textbf{62.44} & \textbf{62.44} & 45.17          & 28.52          & \textbf{46.68}   \\
Baichuan-Audio    & 30.70          & 30.70          & 48.60          & 30.46          & 28.20          & 28.20          & \textbf{54.11} & \textbf{40.66} & 27.47          & 27.47          & 39.96          & 23.94          & 30.49          & 30.49          & 47.15          & \textbf{33.80} & 34.53   \\
GLM-4-Voice       & 26.40          & 26.40          & 10.48          & 19.41          & 19.49          & 19.49          & 9.31           & 39.59          & 32.51          & 32.51          & 9.58           & 15.31          & 28.36          & 28.36          & 10.64          & 19.19          & 21.69   \\
Qwen2-Audio       & 24.76          & 24.78          & 47.67          & 30.67          & 18.14          & 18.26          & 52.76          & 25.48          & 27.47          & 27.50          & 38.58          & 23.78          & 23.09          & 23.15          & 45.69          & 32.92          & 30.29   \\
Baichuan-Omni     & 36.17          & 36.19          & 48.06          & 28.88          & 34.53          & 34.54          & 52.99          & 39.81          & 38.68          & 38.79          & 39.99          & 24.42          & 34.53          & 34.53          & 46.34          & 31.34          & 37.49   \\
MiniCPM-o         & 39.74          & 39.74          & 46.02          & 26.56          & 35.84          & 35.84          & 49.91          & 33.41          & 41.82          & 41.82          & 36.82          & 20.52          & 39.01          & 39.01          & 44.40          & 28.52          & 37.44   \\
Qwen2.5-Omni      & 24.52          & 24.54          & 47.70          & 30.88          & 18.17          & 18.25          & 52.55          & 39.75          & 27.58          & 27.64          & 39.57          & \textbf{24.43} & 22.87          & 22.90          & 45.67          & 33.45          & 31.28   \\
GPT-4o-Audio & 46.31          & 46.31          & \textbf{51.53} & \textbf{31.93} & 45.04          & 45.04          & 53.65          & 38.39          & 48.21          & 48.21          & \textbf{43.16} & 24.27          & 49.10          & 49.10          & \textbf{48.45} & 33.10          & 43.86   
\end{tblr}
\end{table}
\begin{wraptable}{l}{0.5\textwidth}
\centering
\scriptsize
\caption{Performance of the model on the chat task of the RealTalk-CN dataset. The Pipeline represents Whisper-large-v3 + GPT-4o, and the score is the score of GPT-4o-mini-Audio, with a full score of 5. $^*$Note that GPT-4o-Audio-mini does not support speech mode on the assistant side during input.}
\begin{tblr}{
  width = 0.9\linewidth,
  colspec = {Q[300]Q[130]Q[130]Q[120]Q[120]Q[80]}, % 各列宽度略微减小
  cell{2}{6} = {r},
  cell{3}{6} = {r},
  cell{4}{6} = {r},
  cell{5}{6} = {r},
  cell{6}{6} = {r},
  cell{7}{6} = {r},
  cell{8}{6} = {r},
  cell{9}{6} = {r},
  hline{1} = {-}{0.08em},
  hline{2-3,9-10} = {-}{0.05em},
  rowsep = 1pt, % 行间距减小
  colsep = 2pt, % 列间距减小
}
Models            & MD-Col        & MD-Sys        & SD-Col        & SD-Sys        & Avg           \\
Pipline           & \textbf{8.92} & \textbf{9.12} & \textbf{8.84} & \textbf{9.12} & \textbf{9.00} \\
Baichuan-Audio    & 7.44          & 7.80          & 7.79          & 7.68          & 7.67          \\
GLM-4-Voice       & 8.30          & 8.54          & 8.24          & 8.39          & 8.37          \\
Qwen2-Audio       & 7.82          & 8.11          & 7.85          & 8.06          & 7.96          \\
Baichuan-Omni     & 7.32          & 7.51          & 7.34          & 7.72          & 7.47          \\
MiniCPM-o         & 8.22          & 8.41          & 8.19          & 8.33          & 8.29          \\
Qwen2.5-Omni      & 7.83          & 8.14          & 7.78          & 8.04          & 7.95          \\
Gpt-4o-Audio-mini & 8.66$^*$      & 8.79$^*$      & 8.71$^*$      & 8.77$^*$      & 8.73$^*$      
\end{tblr}
\label{table2}
\end{wraptable}

\subsection{Results \& Discussion}
\textbf{Speech disfluency affects slot filling and chat tasks.} In Table~\ref{table1}, on the Colloquial subsets, the performance of most models in the slot filling task dropped significantly, such as the JGA value of the Pipeline method dropped from 31.99 to 26.09, and the Baichuan-Audio dropped from 40.66 to 30.46 on the MD-Col subset, while the intent classification task was not significantly affected. This difference stems from the difference in the performance of the models in processing speech features and semantic information. The intent classification task mainly relies on capturing the core semantics in the user's sentence, even in the case of unfluent speech expression, the model can maintain high accuracy through contextual semantics. However, in the slot filling task, the model needs to accurately identify and extract specific slot information in the continuous speech stream, which relies on the model's ability to capture key information. When there are self-repetitions, grammatical errors, or non-standard expressions in the speech, the model is easily confused in slot boundaries and content extraction, which puts higher requirements on the model. The chat task (shown in Table~\ref{table2}) requires obtaining the core semantics, extracting some key information, and responding after integration. The performance also declines on the Colloquial subsets.

\textbf{The Pipeline method and the end-to-end model have their own advantages and disadvantages}: in the intent classification task shown in Table~\ref{table1} and the chat task shown in Table~\ref{table2}, the Pipeline method performs better than the end-to-end model. This result can be attributed to the fact that in the Pipeline method, after Whisper-large-v3 converts speech into text, GPT-4o is responsible for text semantic understanding and intent classification, making full use of GPT-4o's powerful ability in text understanding, and these two tasks rely more on semantics. In contrast, although the end-to-end model can capture speech and text information directly from the speech input, it often declines in semantic understanding ability. However, in the slot-filling task, the end-to-end model outperforms the Pipeline method. In Table~\ref{table1}, Many models have higher F1 and JGA values than the Pipeline method. Tasks such as slot filling rely more on details in speech, while the end-to-end model can be more adaptable to speech quality and expression clarity (such as disfluency) and is more conducive to capturing detailed information of speech. In addition, after calculating the average of their performance for all tasks, the Pipeline method maintained its leading position, while GPT-4o-Audio-mini was second best, and was generally ahead of other end-to-end models in multiple tasks, indicating that it has stronger capabilities in speech understanding and multimodal feature fusion.
\begin{figure}[tbp]
    \centering
    \includegraphics[width=0.68\textwidth]{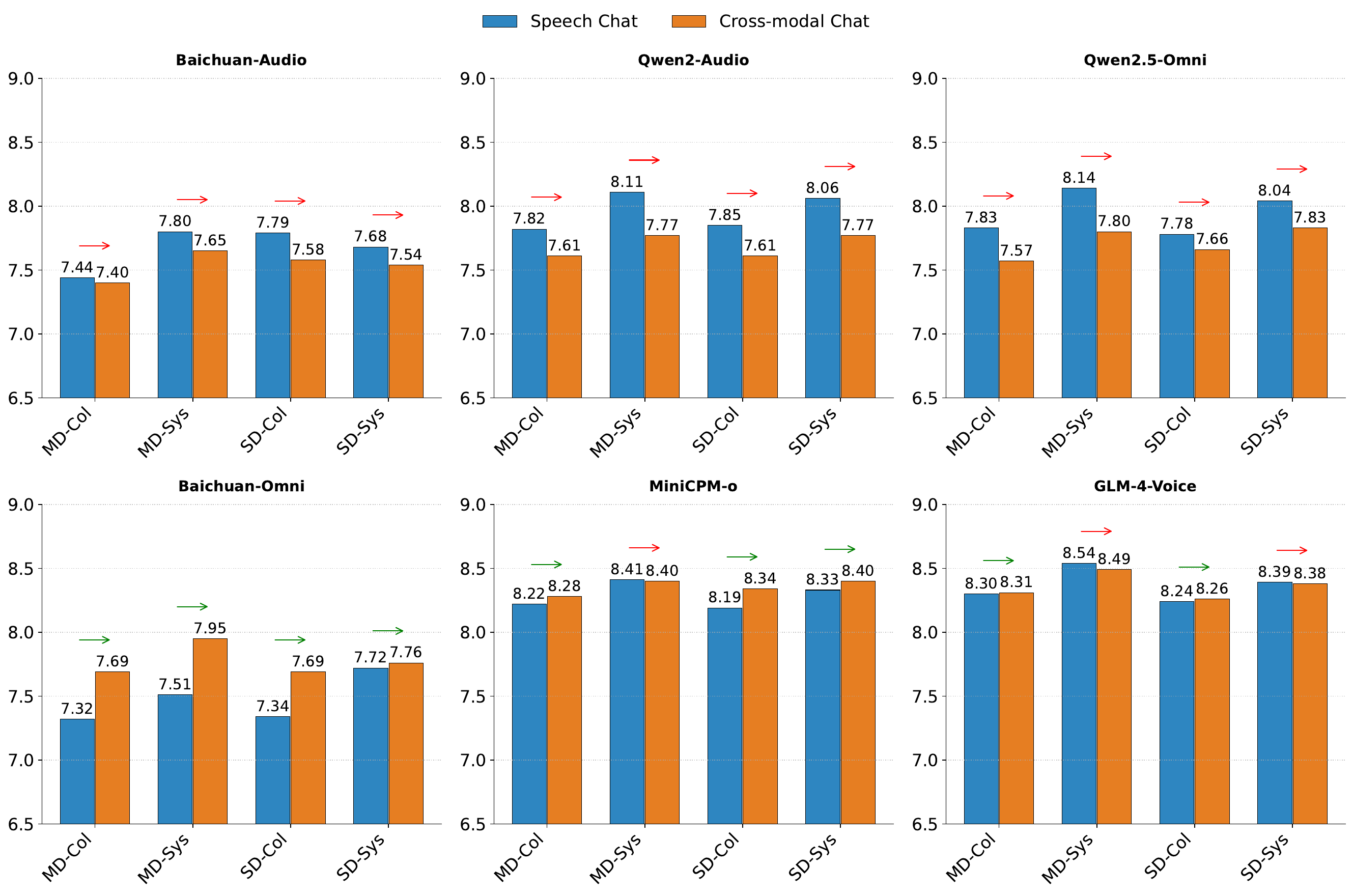}
    \caption{Performance comparison between pure speech chat tasks and Cross-modal chat tasks. The results were analyzed using paired t-tests~\cite{student1908error} (p<0.05), and the tasks with significant differences were Qwen2-Audio, Baichuan-Audio, Baichuan-Omni, and Qwen2.5-Omni, while those with insignificant differences were MiniCPM-o and GLM-4-Voice. The detailed process can be found in Appendix C.3.}
    \label{fig:Modal_results}
\end{figure}

\textbf{Multi-domain complexity mainly affects intent classification capabilities} On the Multi-Domain subset, the performance of intent classification tasks is significantly lower than that of the Single-Domain subset. For example, in Table~\ref{table1}, compared with SD-Col, the PANDA of intent classification of MD-Col is generally reduced by 2-5 points, while the performance of slot filling tasks is not significantly affected. This difference reflects the limitations of the model in dealing with semantic diversity and context switching. The intent classification task essentially relies on the model to correctly classify user intent in the semantic space, and multi-domain scenarios involve multiple tasks and contexts, so the model needs to have stronger cross-domain semantic generalization capabilities. However, current end-to-end speech models often lack semantic representation and context adaptability when faced with domain switching, and may mistakenly confuse the semantics of different domains, thus affecting the accuracy of intent classification. In contrast, the slot filling task performs more stably in multi-domain scenarios because it relies on the recognition of specific slots. The model only needs to recognize predefined slot information, and domain changes have little impact on the definition of these slots.

\textbf{Performance Divergence in Cross-Modal Chat Tasks.} Figure~\ref{fig:Modal_results} illustrates the varied performance of speech foundation models across pure speech-based chat tasks and cross-modal chat tasks, revealing distinct model behaviors. The first category comprises models exhibiting performance degradation, including Baichuan-Audio, Qwen2-Audio, and Qwen2.5-Omni. These models show consistent metric declines in cross-modal scenarios, exemplified by Qwen2-Audio's MD-Col score decreasing from 7.82 to 7.61. Through a detailed case study, we found that the model did have some problems when integrating and responding to heterogeneous modal information, including forgetting the key information and semantics of the previous context of different modalities, reduced quality of response richness, and response text degeneration. The detailed case can be found in Appendix E.1. The second category features models maintaining stable performance, including GLM-4-Voice and MiniCPM-o. In particular, the third category contains models that achieve performance improvements. Baichuan-Omni shows a significant MD-Col score increase from 7.32 to 7.69, which suggests that this model benefits from text-modality substitutions in dialogue history.

\begin{figure}[htbp]
    \centering
    \includegraphics[width=0.71\textwidth]{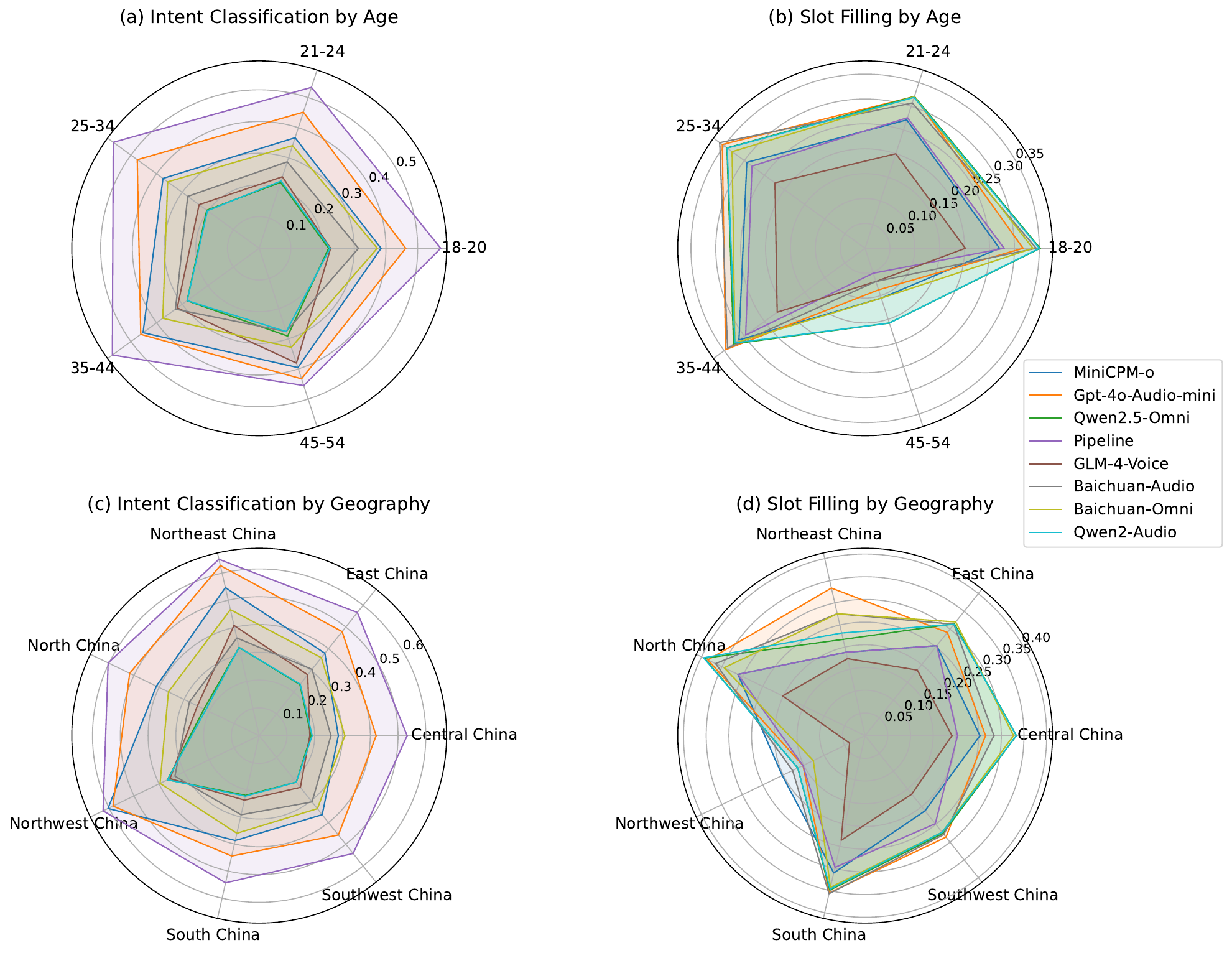}
    \caption{Radar chart showing the impact of speaker's age and geography on dialogue intent classification and slot filling tasks. The result is the average performance of the model on the entire RealTalk-CN dataset, grouped by age and geography (which refers to ancestral origin). We performed an overall Kruskal-Wallis H test~\cite{kruskal1952ranks} (p<0.05) and a comparative Mann-Whitney U test~\cite{mann1947test} (p<0.05) for significance, and calculated Cohen's d effect size~\cite{cohen1988power}. Details are available in Appendix C.2.}
    \label{fig:Speaker_verification}
\end{figure}

\subsection{Speaker verification}
Speaker-related attributes such as age~\cite{kulkarni2024bias}~\cite{chen2025seniortalk} and regional accent~\cite{chen2024voicebench} significantly influence speech model performance. Existing studies~\cite{kulkarni2024bias}~\cite{chen2025seniortalk} demonstrate that current speech models exhibit notable performance degradation when processing elderly users' speech due to age-related vocal deterioration~\cite{fraser2015alzheimer}. While~\cite{chen2024voicebench} investigated global English accent variations' impact on speech LLMs, comparable research remains scarce for Chinese speech LLMs, particularly in task-oriented dialogue domains. Given China's vast geographical distribution with diverse Mandarin accents and broad age demographics, we systematically analyze age and regional accent effects on speech-based LLMs in Chinese task-oriented dialogue Scenarios, with results shown in Figure~\ref{fig:Speaker_verification}.

From a task perspective, models exhibit differential sensitivity to speaker characteristics during intent classification and slot filling. Fewer models showed statistically significant differences in intent classification, particularly across age groups, except for notable variations in Northwest and Northeast China's geographical distribution. In contrast, slot filling demonstrated pronounced susceptibility to both age and regional factors (p<0.05), reinforcing our findings in Section 5.3 that fine-grained semantic parsing tasks are more vulnerable to speech variability.

Demographic analysis revealed that slot filling performance for 45-54-year-old users decreased significantly with strongly negative Cohen's d values, confirming the compounded challenges of vocal aging and distributional mismatch with training data. Regional comparisons further indicated superior performance for South and North China users compared to Southwest or Northwest regions, suggesting standard Mandarin proximity in training data better serves linguistically central regions, while stronger dialectal features in peripheral areas degrade model generalization.

Model adaptation capabilities varied substantially. In slot filling, Baichuan-Audio and GPT-4-Audio-mini excelled with mainstream speakers but suffered >20\% performance degradation for elderly users, indicating limited robustness to vocal aging. Conversely, Qwen models maintained consistent performance across age groups. For regional adaptation, MiniCPM-o, GPT-4-Audio-mini, and Qwen demonstrated superior cross-regional generalization. These findings highlight critical directions for enhancing model fairness—improving stability across diverse user demographics.

\section{Conclusion}

In this paper, we introduce RealTalk-CN, the first large-scale Chinese speech-text dual-modal dialogue benchmark that comprehensively captures speech disfluencies, diverse speaker characteristics, and cross-modal interactions. Our evaluations demonstrate the dataset’s effectiveness in benchmarking models on speech robustness, speaker adaptation, and cross-modal consistency. The proposed cross-modal chat task further reveals models' limitations in handling dynamic modality switching. RealTalk-CN sets a new standard for Chinese multimodal dialogue research, providing a critical resource for advancing speech-based language models.
\bibliographystyle{unsrtnat}
\bibliography{ref}

\end{document}